\documentclass[10pt,twocolumn,letterpaper]{article}

\usepackage[pagenumbers]{cvpr} % To force page numbers, e.g. for an arXiv version

\definecolor{cvprblue}{rgb}{0.21,0.49,0.74}
\usepackage[pagebackref,breaklinks,colorlinks,allcolors=cvprblue]{hyperref}

\graphicspath{{./figures/}}

\usepackage{color}
\usepackage{colortbl}
\usepackage{xcolor}
\usepackage{tikz}
\definecolor{gold}{HTML}{BD820B}%{EEAD0E}
\definecolor{silver}{HTML}{909090}%{C0C0C0}
\definecolor{bronze}{HTML}{9A5F26}%{CD7F32}
\definecolor{lgray}{gray}{0.95}

\newcommand*\circledd[1]{\tikz[baseline=(char.base)]{
            \node[shape=circle,draw,inner sep=0.15pt] (char) {#1};}}     
            
\newcommand{\first}[1]{%
    {#1\raisebox{0.8pt}{\footnotesize \color{gold} \circledd{1}}}%
}
\newcommand{\second}[1]{%
    {#1\raisebox{0.8pt}{\footnotesize \color{silver} \circledd{2}}}%
}
\newcommand{\third}[1]{%
    {#1\raisebox{0.8pt}{\footnotesize \color{bronze} \circledd{3}}}%
}

\newcommand\blfootnote[1]{%
  \begingroup
  \renewcommand\thefootnote{}\footnote{#1}%
  \addtocounter{footnote}{-1}%
  \endgroup
}

\title{A Distractor-Aware Memory for Visual Object Tracking with SAM2}

\author{Jovana Videnovic$^\ast$, Alan Lukezic$^\ast$, Matej Kristan \\
\small{Faculty of Computer and Information Science, University of Ljubljana, Slovenia} \\
{\tt\small jv8043@student.uni-lj.si, \{alan.lukezic, matej.kristan\}@fri.uni-lj.si}
}

\begin{document}
\maketitle

\blfootnote{$^\ast$ The authors contributed equally.}

\begin{abstract}
Memory-based trackers are video object segmentation methods that
form the target model by concatenating recently tracked frames into a memory buffer and localize the target by attending the current image to the buffered frames. 
While already achieving top performance on many benchmarks, it was the recent release of SAM2 that placed memory-based trackers into focus of the visual object tracking community. 
Nevertheless, modern trackers still struggle in the presence of distractors. We argue that a more sophisticated memory model is required, and propose a new distractor-aware memory model for SAM2 and an introspection-based update strategy that jointly addresses the segmentation accuracy as well as tracking robustness. The resulting tracker is denoted as SAM2.1++.
We also propose a new distractor-distilled DiDi dataset to study the distractor problem better. SAM2.1++ outperforms SAM2.1 and related SAM memory extensions on seven benchmarks and sets a solid new state-of-the-art on six of them.
The code and the new dataset will be available on \url{https://github.com/jovanavidenovic/DAM4SAM}.
\end{abstract}    
\section{Introduction}  \label{sec:intro}

\begin{figure}[t]
  \centering
  \includegraphics[width=\linewidth]{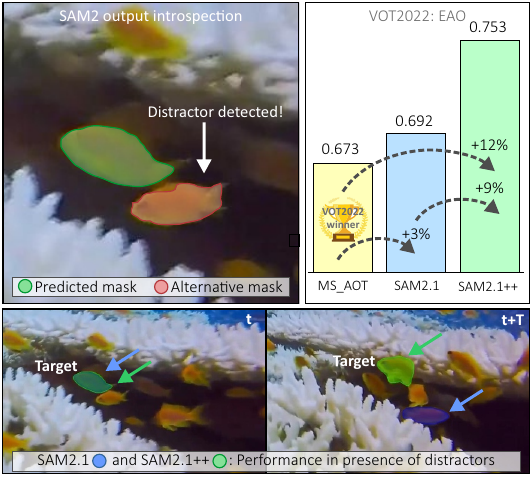} 
  \caption{SAM2.1++ distractor-aware memory (DAM) update is triggered by the divergence between the predicted and the alternative masks (top left). 
  This resolves the visual ambiguity and increases tracking robustness (bottom).
  DAM leads to a significant performance boost, setting a new sota on VOT2022 (top-right).}
\label{fig:fig1}
\end{figure}

General visual object tracking is a classical computer vision problem that considers the
localization of an arbitrary target in the video, given a single supervised training example in the first frame. The major source of tracking failures are so-called distractors, i.e., image regions that are difficult to distinguish from the tracked object, given the available target model (see Figure~\ref{fig:fig1}). 
These can be nearby objects similar to the tracked target (\textit{external distractors}) or similar regions on the object when tracking only a part of the object (\textit{internal distractors}).
When the target leaves and re-enters the field of view, the external distractors become particularly challenging.

Various approaches have been proposed to reduce the visual ambiguity caused by distractors. 
These include learning discriminative features~\cite{siamfc_eccvw2016,stark_iccv21,transt_cvpr2021,mixformer_cvpr2022, seqtrack} or explicitly modeling the foreground-background by dedicated modules~\cite{d3s_tpami,odtrack,cutie_cvpr2024,ditra_ijcv2024}.
An emerging paradigm, already positioned at the top of the major benchmarks~\cite{vot2022, vots2023, vots2024}, are memory-based frameworks, which localize the target by pixel association with the past tracked frames~\cite{ms_aot,xmem_eccv2022,rmem_cvpr2024}. 

The memory-based methods construct the target model by concatenating sequences of entire images with the segmented target, thus implicitly encoding the present distractors.
~\cite{rmem_cvpr2024} argue that the visual redundancy in the large memory leads to reduced localization capability due to the nature of cross-attention. They show that limiting the memory to the most recent frames and temporally time-stamping them in fact improves tracking.
This paradigm was further verified by the recent tracking foundation model SAM2~\cite{sam2}, which sets a solid state-of-the-art across several video segmentation and tracking benchmarks~\cite{lvos,ytvos,  mose,vots2024, davis17}. 

We argue that while the recent target appearances in the memory are required for accurate segmentation, another type of memory is required to distinguish the target from challenging distractors.
To support this claim, we propose a new distractor-aware memory (DAM) and update mechanism for SAM2.
The new memory is divided by its tracking functionality into two parts: the recent appearances memory (RAM) and distractor-resolving memory (DRM). 
While RAM contains the recent target appearances sampled at regular intervals, DRM contains anchor frames that help discriminate the target from critical distractors. 
A novel DRM updating mechanism is proposed that exploits the information of SAM2 output, which has been ignored so far by the tracking research.

In addition, we observe that standard benchmarks contain many sequences, which are no longer considered challenging by modern standards. The high performance on these sequences overwhelms the total score, saturates the benchmarks, and does not properly expose the tracking advances. 
To address this, we semi-automatically distill several benchmarks into a distractor-distilled tracking dataset (DiDi). 

In summary, our main contribution is the new distractor-aware memory DAM for SAM2 and the updating strategy, resulting in SAM2++. 
To the best of our knowledge, this is the first memory formulation that divides and updates the memory with respect to its function in tracking. Our secondary contribution is the new DiDi dataset that more clearly exposes tracking advances in the presence of distractors.
With no additional training, SAM2.1++ substantially outperforms SAM2.1 in robustness on several standard bounding box and segmentation tracking benchmarks, including the new DiDi dataset, and sets a new state-of-the-art in visual object tracking.

\section{Related Work}  \label{sec:related_work}

Transformers are currently the dominant methodology in visual object tracking~\cite{vot2022} and can be largely categorized into classification- and regression-based~\cite{transt_cvpr2021}, corner prediction-based~\cite{stark_iccv21}, and sequence-learning-based trackers~\cite{seqtrack}. 

The recent top-performing tracking frameworks are inspired by the video object segmentation methods based on memory networks~\cite{xmem_eccv2022, ms_aot,cutie_cvpr2024,sam2, rmem_cvpr2024}. These methods embed predictions from past frames into memory, therefore extending contextual information beyond just the initial or the previous frame.
The attention mechanism is typically used to link frame representations stored in the memory with features extracted from the current frame.
In the initial methods like~\cite{ms_aot}, the arriving frames were continually added to the memory. This led to theoretically unbounded increase in computational complexity and GPU memory.

This issue was addressed in~\cite{xmem_eccv2022, cutie_cvpr2024} by using multiple memory storages and efficient compression schemes to capture different temporal contexts, enhancing  performance on long-term videos. 
Alternatively,~\cite{rmem_cvpr2024} proposed to restrict the memory to the most recent frames with temporal stamping, which led to improved localization. 
The principle of the restricted memory is followed by the SAM2~\cite{sam2} foundation model, which stores last 6 frames and the initial frame in the memory. 
Recently, SAM2Long~\cite{sam2long} proposed a training-free method to enhance the performance of SAM2~\cite{sam2} on long-term sequences by determining the optimal trajectory from multiple segmentation pathways using a constrained tree search.

Most of the existing tracking methods do not explicitly address tracking in the presence of distractors, even though distractors are a major source of tracking failures. 
Discriminative (deep) correlation filters~\cite{danelljan_dimp_iccv19,tomp_cvpr2022} are theoretically suitable to handle distractors but are in practice outperformed by the modern transformer-based trackers. 
However, there have been some recent attempts to address distractors.
KeepTrack~\cite{keeptrack} casts the problem as a multi-target tracking setup, where it identifies target candidates and potential distractors, which are then associated with previously propagated identities using a learned association network. However, the method relies on accurate detection and cannot address internal distractors in practice. 
In~\cite{ditra_ijcv2024} target localization accuracy and robustness are treated as two distinct tasks, which is demonstrated to be beneficial in situations with distractors. 
Despite explicit distractor handling mechanism, these methods lead to complicated architectures and cannot fully exploit the learning potential of modern frameworks. 
In contrast, memory-based methods~\cite{xmem_eccv2022, cutie_cvpr2024, rmem_cvpr2024, sam2} have the capacity to implicitly handle distractors in an elegant way, since they store entire images and apply a learnable localization by segmentation. However, the existing memory management methods are not designed to effectively handle the distractors. 

\section{Distractor-aware memory for SAM2}  \label{sec:method}

This section describes the new DAM memory model for SAM2. Section~\ref{sec:sam2} briefly outlines the SAM2 architecture, while the new model is described in Section~\ref{sec:memory_management}.

%-------------------------------------------------------------------------
\subsection{SAM2 preliminaries}  \label{sec:sam2}

SAM2 extends the Segment Anything Model (SAM)~\cite{sam_iccv2023}, originally developed for interactive class-agnostic image segmentation, to video segmentation. It consists of four main components: (i) image encoder, (ii) prompt encoder, (iii) memory bank, and (iv) mask decoder.

The image encoder applies ViT Hiera\footnote{ViT-L version is used in all our experiments.} 
backbone~\cite{hiera} to embed the input image.
Interactive inputs (e.g., positive/negative clicks) are absorbed by the prompt encoder and used for output mask refinement, however, note that these are not applicable in the general object tracking setup. 
The memory bank consists of the encoded initialization frame with a user-provided segmentation mask and six recent frames with segmentation masks generated by the tracking output. Temporal encodings are applied to the six recent frames to encode the frame order, while such encoding is not applied to the initialization frame to indicate its unique property of being a single supervised training example and thus serves as a sort of target prior model.

The memory bank transfers pixel-wise labels onto the current image by attending the features in the current frame to all memory frames, producing memory-conditioned features. The features are then decoded by the mask decoder, which predicts three output masks along with the IoU prediction for each. The mask with the highest IoU is chosen as the tracking output.

SAM2 applies a variant of the memory management proposed in~\cite{rmem_cvpr2024}. The initialization frame is always kept in the memory, while the the six recent frames are updated at every new frame by a first-in-first-out protocol. The memory and the management mechanism are visualized in  Figure~\ref{fig:memory}, while the reader is referred to~\cite{sam2} for other details.

\subsection{Distractor-aware memory -- DAM}  \label{sec:memory_management}

Related works~\cite{ms_aot,xmem_eccv2022,rmem_cvpr2024,sam2} have clearly demonstrated the importance of the most recent frames, which are required to address target appearance changes and ensure accurate segmentation. However, a different type of frames is required to prevent drifting in the presence of critical distractors and for reliable target re-detection.

\begin{figure*}[ht]
  \centering
  \includegraphics[width=\linewidth]{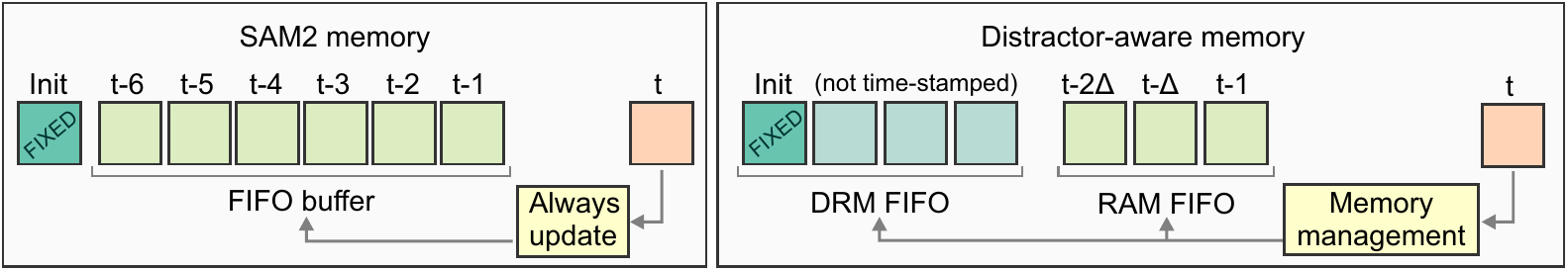} 
  \caption{Overview of the SAM2 memory and the proposed Distractor-Aware Memory (DAM), which splits the model into
  {\it Recent Appearance Memory} (RAM) and 
  {\it Distractor Resolving Memory} (DRM) and updates them by a new memory management protocol.}
\label{fig:memory}
\end{figure*}

We propose to compose the memory with respect to its function during tracking into (i) \textit{recent appearance memory} (RAM) and (ii) \textit{distractor resolving memory} (DRM). 
RAM and DRM together form the distractor-aware memory (DAM), visualized in Figure~\ref{fig:memory}. 
The function of RAM is to ensure segmentation accuracy in the considered frame, thus we design it akin to the current SAM2~\cite{sam2} memory. It is composed of a FIFO buffer with 
${1 \over 2}N_\mathrm{DAM}=3$
slots, contains the most recent target appearances, and applies temporal encoding to identify temporally more relevant frames for the task. 

On the other hand, DRM serves for ensuring tracking robustness and re-detection. It should contain accurately segmented frames with critical recent distractors, including the initialization frame. 
It is thus composed of a slot reserved for the initialization frame and a FIFO buffer with ${1 \over 2}N_\mathrm{DAM}=3$ anchor frames updated during 
tracking\footnote{At tracker initialization, RAM occupies all $N_\mathrm{DAM}$ slots and drops to ${1 \over 2}N_\mathrm{DAM}$ as DRM entries arrive, to fully exploit the available capacity.}.
Since the purpose of DRM is to encode critical information for resolving distractors, it does not apply temporal encoding. Note that the pre-trained SAM2 already contains the building blocks to implement the proposed memory structure. 

\subsubsection{RAM management protocol}  \label{sec:memory_ram}

A crucial element of the memory-based methods is the memory management protocol. 
To efficiently exploit the available memory slots, the memory should not be updated at every frame, since the consecutive frames are highly correlated. 
In fact,~\cite{rmem_cvpr2024} argue that visual redundancy in memory should be avoided in attention-based localization. RAM is thus updated every $\Delta=5$ frames and includes the most recent frame since it is the most relevant for accurate target segmentation in the considered frame.

SAM2~\cite{sam2} updates the memory at every frame, including when the target is absent. However, even for a very short occlusion, the memory will quickly fill up with frames without the target, which reduces the target appearance diversity in the model, leading to a reduced segmentation accuracy upon target re-appearance. Furthermore, failing to re-detect the target leads to incorrectly updating the memory by an empty mask, which may cause error accumulation and ultimately re-detection failure. We thus propose to not update RAM when the target is not present, i.e., when the predicted target mask is empty.

\subsubsection{DRM management protocol}  \label{sec:memory_drm}

DRM inherits the initial update rules from RAM, i.e., update only when the target is present and at least $\Delta=5$ frames have passed since the last update.
It considers an additional rule to identify anchor frames containing critical distractors. 
In particular, drifting to a distractor may be avoided by including a past temporally nearby frame with this distractor accurately segmented as the background. 
Recall that SAM2 predicts three output masks and selects the one with the highest predicted IoU (Section~\ref{sec:sam2}), which means we can consider it as a multi-hypothesis prediction model.
Our preliminary study showed that in the frames before the failure occurs, SAM2 in fact detects such distractors in the alternative two predicted output masks (see Figure~\ref{fig:fig1}). We thus propose a simple
anchor frame detection mechanism based on determining hypothesis divergence between the output and alternative masks.

A bounding box is fitted to the output mask and to the union of the output mask and the largest connected component in the alternative mask. 
If the ratio between the area of the two bounding boxes drops below $\theta_\mathrm{anc}=0.7$, the current frame is considered as a potential candidate to update DRM. 
Note that updating with a grossly incorrectly segmented target would lead to memory corruption and eventual tracking failure. 
We thus trigger the DRM update only during sufficiently stable tracking periods, i.e., if the predicted IoU score from SAM2 exceeds a threshold $\theta_\mathrm{IoU}=0.8$ and if the mask area is within $\theta_\mathrm{area}=20\%$ of the median area in the last $\theta_{M}=10$ frames. 
Note that SAM2.1++ is insensitive to the exact value of these parameters. 

%*************************************************************************
\section{A distractor-distilled dataset}  \label{sec:disd_dataset}

While benchmarks played a major role in the recent visual object tracking breakthroughs, we note that many of these contain sequences, which are no longer considered challenging by modern standards. In fact, most of the modern trackers obtain high performance on these sequences, which overwhelms the total score and under-represents the improvements on challenging situations.
To facilitate the tracking performance analysis of the designs proposed in this paper, we semi-automatically distill several benchmarks into a distractor-distilled tracking dataset (DiDi).

We considered validation and test sequences of the major tracking benchmarks, which are known for high-quality annotation, i.e., GoT-10k~\cite{got10k}, LaSOT~\cite{lasot_cvpr19}, UTB180~\cite{utb180}, VOT-ST2020 and VOT-LT2020~\cite{kristan_vot2020}, and VOT-ST2022 and VOT-LT2022~\cite{vot2022}. This gave us a pool of 808 sequences. 
A sequence was selected for the DiDi dataset if at least one-third of the frames passed the distractor presence criterion described next.

A frame was classified as containing non-negligible distractors if it contained a large enough region visually similar to the target. This criterion should be independent from tracker localization method, yet should reflect the power of the modern backbones. We thus encoded the image by DINO2~\cite{dino2} and for each pixel in the feature space computed the distractor score as the average cosine distance to the features within the ground truth target region. We then computed the ratio between the number of pixels outside and inside the target region that exceeded the average of the scores computed within the target region. If the ratio exceeded 0.5, we considered the frame as containing non-negligible distractors.

Using the aforementioned protocol, we finally obtained 180 sequences with an average sequence length of 1.5k frames (274,882 frames in total). Each sequence contains a single target annotated by an axis-aligned bounding box. In addition, we manually segmented the initial frames to allow the initialization of segmentation-based trackers. Figure~\ref{fig:dataset} shows frames from the proposed DiDi dataset. Please see the supplementary material for additional information.

\begin{figure*}[ht]
  \centering
  \includegraphics[width=\linewidth]{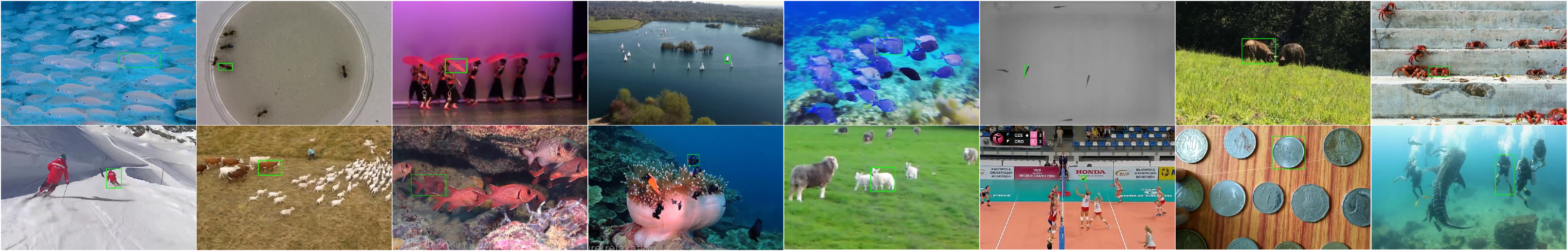} 
  \caption{Example frames from the DiDi dataset showing challenging distractors. Targets are denoted by green bounding boxes.}
\label{fig:dataset}
\end{figure*}

\section{Experiments}  \label{sec:experiments}

The proposed DAM for SAM2 memory model is rigorously analyzed. Section~\ref{sec:exp-justification} reports a battery of experiments to justify the design choices. Section~\ref{sec:exp-disd} compares the SAM2.1 extension with DAM memory with the state-of-the-art on the DiDi dataset. 
Detailed analysis on the challenging VOT tracking-by-segmentation benchmarks is reported in Section~\ref{sec:exp-vot}, while a comparison on the standard bounding-box tracking benchmarks is reported in Section~\ref{sec:exp-bb}.

%-------------------------------------------------------------------------
\subsection{Architecture justification}  \label{sec:exp-justification}

\begin{figure}[ht]
  \centering
  \includegraphics[width=\linewidth]{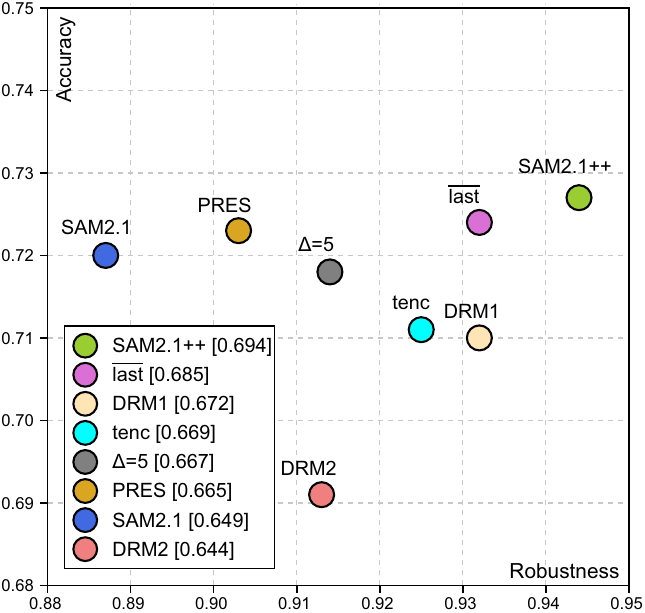} 
  \caption{
  Accuracy-robustness plot on DiDi for the ablated versions of SAM2.1++. The tracking quality is given at each label.
  }
\label{fig:ar_ablation}
\end{figure}

The design choices of the proposed distractor-aware memory and the management protocol from Section~\ref{sec:memory_management} are validated on the DiDi dataset from Section~\ref{sec:disd_dataset}. 
We compute the VOTS~\cite{vots2023} measures since they simultaneously account for short-term as well as long-term tracking performance. 
Performance is summarized by the tracking quality Q score and two auxiliary measures: robustness (i.e., portion of successfully tracked frames) and accuracy (i.e., the average IoU between the prediction and ground truth during successful tracking). 
Results are shown in Table~\ref{tab:architecture_justification} and in AR plot in Figure~\ref{fig:ar_ablation}.

We first verify the argument from Section~\ref{sec:memory_ram}, that updating with frames without a target present causes memory degradation. 
We thus extend SAM2.1, with updating only when the predicted mask is not empty (denoted as SAM2.1$_{\mathrm{PRES}}$). 
SAM2.1$_{\mathrm{PRES}}$ increases the tracking quality Q by 2.5\%, primarily by improved robustness, which justifies our claim.

We next verify the assumption (also claimed in~\cite{rmem_cvpr2024}) that frequent updates reduce tracking robustness due to highly correlated information stored in memory. 
We reduce the update rate in SAM2.1$_{\mathrm{PRES}}$ to every 5th frame (SAM2.1$_{\Delta=5}$). 
This negligibly improves Q, but does increase the robustness by 1.2\%, which supports the claim. 
We did not observe further performance improvement with increasing $\Delta$.

Finally, we focus on the distractor-resolving memory (DRM) part of our proposed memory in Section~\ref{sec:memory_ram}, that we claim is responsible for the tracking robustness in the presence of distractors. 
Recall that DRM is updated when a distractor is detected and under the condition that tracking is reliable -- we first test the influence of these two conditions independently. 
We thus extend SAM2.1$_{\Delta=5}$ with the new DAM memory and update the DRM part only during reliable tracking periods (SAM2.1$_{DRM1}$). The tracking accuracy improves a little, with robustness increase of 2\%. 
Alternatively, we change the rule to update only when the distractor is detected (SAM2.1$_{DRM2}$). While robustness remains unchanged w.r.t. SAM2.1$_{\Delta=5}$, the accuracy in fact drops. 
This is expected since distractor detection may be triggered also by the error in the target segmentation, which gets amplified by the update. To verify this, we next apply both our proposed update DRM rules, arriving to the proposed DAM with SAM2.1 (SAM2.1++ for short). 
Compared to SAM2.1$_{\Delta=5}$, we observe substantial improvement in tracking quality Q (4\%), primarily due to 3.3\% boost in robustness as well as 1.3\% boost in accuracy, taking the top-right position in the AR plot (Figure~\ref{fig:ar_ablation}) among all variants.
This conclusively verifies that DRM should be updated with distractor detected only if tracking is sufficiently reliable.

\begin{table}[t]
\centering
\caption{SAM2.1++ architecture justification on DiDi dataset.}
\label{tab:architecture_justification}
\begin{tabular}{lccc}
\toprule
  & Quality & Accuracy & Robustness \\
\midrule
SAM2.1        & 0.649 & 0.720 & 0.887 \\
SAM2.1$_{\mathrm{PRES}}$       & 0.665 & 0.723 & 0.903 \\
SAM2.1$_{\Delta=5}$ & 0.667 & 0.718 & 0.914 \\
SAM2.1$_{\mathrm{DRM1}}$   & 0.672 & 0.710 & 0.932 \\ % !! SAM2.1-STLT
SAM2.1$_{\mathrm{DRM2}}$  & 0.644 & 0.691 & 0.913 \\ % !! SAM2.1-distr
SAM2.1++      & 0.694 & 0.727 & 0.944 \\
\midrule
DRM$_{\mathrm{tenc}}$       & 0.669 & 0.711 & 0.925 \\
RAM$_{\mathrm{\overline{last}}}$       & 0.685 & 0.724 & 0.932 \\
\bottomrule
\end{tabular}
\end{table}

In Section~\ref{sec:memory_ram} we claim that DRM part of the memory should not be time-stamped, since the frame influence on the distractor resolving in the current frame should not be biased by the temporal proximity and should serve as a time-less prior. 
To test this claim, we modify SAM2.1++ by using temporal encodings in DRM (except for the initialization frame) -- we denote it as DRM$_{\mathrm{tenc}}$. 
The tracking quality drops by 3.6\%, which confirms the claim. 
We further inspect the updating regime in RAM, which always includes the most recent frame, but updates the memory slots every 5th frame. 
SAM2.1++ is modified to update all RAM slots at every 5th frame (RAM$_{\mathrm{\overline{last}}}$). This results in a slight tracking quality decrease (1.3\%), which indicates that including the most recent frame in RAM is indeed beneficial, but not critical.

%-------------------------------------------------------------------------
\subsection{SoTa comparison on DiDi}  \label{sec:exp-disd}

SAM2.1++ is compared on the DiDi dataset with the recent state-of-the-art trackers
TransT~\cite{transt_cvpr2021}, SeqTrack~\cite{seqtrack}, AQATrack~\cite{aqatrack} and AOT~\cite{ms_aot}
as well as trackers with explicit distractor handling mechanisms: 
KeepTrack~\cite{keeptrack}, Cutie~\cite{cutie_cvpr2024} and ODTrack~\cite{odtrack}. 
For completion, we include the most recent, yet unpublished, tracker with improved long-term memory update for SAM2, named SAM2.1Long~\cite{sam2long}.

Results in Table~\ref{tab:dida} reveal the advantages of the trackers with an explicit distractor-handling mechanism over the other trackers. Consider two similarly complex recent trackers SeqTrack and ODTrack, which are both based on the ViT-L backbone. 
On classical benchmarks like LaSoT, LaSoT$_{\mathrm{ext}}$ and GoT10k, ODTrack outperforms SeqTrack by 2\%, 6\% and 4\%, respectively (see Table~\ref{tab:bbox}).
However, the performance gap increases to 15\% on DiDi (Table~\ref{tab:dida}), which confirms that distractors are indeed a major challenge for modern trackers and that DiDi has a unique ability to emphasize the tracking capability under these conditions and to expose tracking design weaknesses.

Focusing on the evaluation of the proposed tracker, SAM2.1++ outperforms all trackers -- the standard state-of-the-art trackers as well as trackers with explicit distractor handling mechanism. 
In particular, SAM2.1++ outperforms state-of-the-art distractor-aware trackers ODTrack and Cutie,  by 14\% and 21\%, respectively. 

We compare the proposed SAM2.1++ with the concurrent, unpublished work SAMURAI~\cite{samurai}. 
SAMURAI is also built on top of SAM2.1~\cite{sam2}, focuses on handling distractors and improves memory management by incorporating motion cues into memory selection and mask refinement process. In this respect, the work is closely related to ours. 
Results show that SAM2.1++ outperforms SAMURAI in tracking quality by 2\% on DiDi, mostly due to the higher robustness (i.e., SAM2.1++ tracks longer than SAMURAI). 
This result demonstrates the superiority of our new DAM memory and the management protocol for distractor handling, which is also less complex than the concurent counterpart proposed in SAMURAI.

Compared to another unpublished tracker with the alternative memory design SAM2.1Long~\cite{sam2long}, SAM2.1++ outperforms it by a healthy 7\% tracking quality boost, indicating superiority of our proposed memory. The results reveal that the major source of the performance boost is the SAM2.1++ tracking robustness, which means it fails less often and thus much better handles distractors.
In fact, a close inspection of the results shows that SAM2.1Long performs on par with the baseline SAM2.1, which indicates that the long-term memory update mechanism presented in~\cite{sam2long} does not improve performance in the presence of distractors. 
Finally, comparing SAM2.1++ to the baseline SAM2.1 reveals 7\% boost in tracking quality, again, attributed mostly due to the improved robustness (6\%). 

\begin{figure}[ht]
  \centering
  \includegraphics[width=\linewidth]{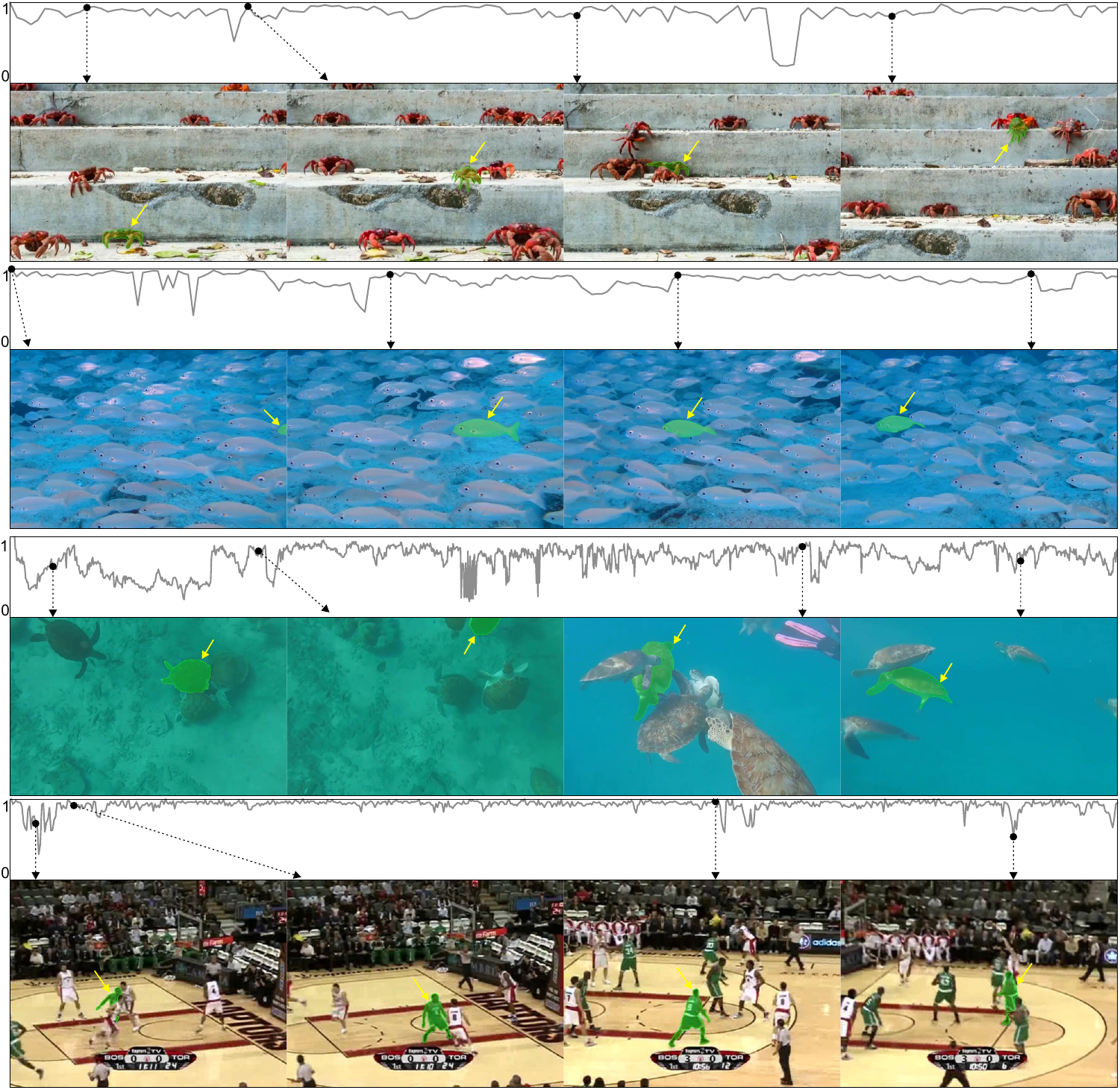} 
  \caption{SAM2.1++ qualitative results on the DiDi dataset with predicted masks shown in green, and tracked objects denoted by arrows. 
  Per-frame overlaps are shown above the figures to indicate failure-free tracking over the entire sequence.}
\label{fig:sam2pp_qualitative}
\end{figure}

These results validate the benefits of the proposed DAM memory and its management protocol in handling challenging distractors. 
Qualitative results of tracking and segmentation with SAM2.1++ on DiDi (Figure~\ref{fig:sam2pp_qualitative}) further demonstrate the remarkable tracking capability in the presence of challenging distractors.

\begin{table}[ht]
\centering
\caption{State-of-the-art comparison on DiDi dataset.}
\label{tab:dida}
\begin{tabular}{llll}
\toprule
  & Quality & Accuracy & Robustness \\
\midrule
SAMURAI \small{~\cite{samurai}}    & 0.680 \second{} & 0.722 \third{} & 0.930 \second{} \\
SAM2.1Long \small{~\cite{sam2long}}  & 0.646 & 0.719 & 0.883  \\
ODTrack \small{~\cite{odtrack}}    & 0.608 & {\bf 0.740} \first{} & 0.809  \\
Cutie  \small{~\cite{cutie_cvpr2024}}     & 0.575 & 0.704 & 0.776  \\
AOT \small{~\cite{ms_aot}}        & 0.541 & 0.622 & 0.852  \\
AQATrack \small{~\cite{aqatrack}}    & 0.535 & 0.693 & 0.753  \\
SeqTrack \small{~\cite{seqtrack}}   & 0.529 & 0.714 & 0.718  \\
KeepTrack \small{~\cite{keeptrack}}  & 0.502 & 0.646 & 0.748  \\
TransT  \small{~\cite{transt_cvpr2021}}    & 0.465 & 0.669 & 0.678  \\
\midrule
SAM2.1 \small{~\cite{sam2}}       & 0.649 \third{} & 0.720 & 0.887 \third{} \\
SAM2.1++      & {\bf 0.694} \first{} & 0.727 \second{} & {\bf 0.944} \first{} \\
\bottomrule
\end{tabular}
\end{table}

%-------------------------------------------------------------------------
\subsection{SoTa comparison on VOT benchmarks}  \label{sec:exp-vot}

The VOT initiative~\cite{kristan_vot_tpami2016} is the major tracking initiative, providing challenging datasets for their yearly challenges. 
In contrast to most of the tracking benchmarks, the targets are annotated by segmentation masks, which allows more accurate evaluation of segmentation trackers, compared to the classic bounding-box benchmarks. 
In this paper, we include two recent single-target challenges: VOT2020~\cite{kristan_vot2020} and VOT2022~\cite{vot2022}, as well as the most recent multi-target challenge VOTS2024~\cite{vots2024}. 

{\bf The VOT2020 benchmark}~\cite{kristan_vot2020} consists of 60 challenging sequences, while trackers are run using anchor-based protocol~\cite{kristan_vot2020} to maximally utilize each sequence. 
Tracking performance is measured by the accuracy and robustness, summarized by the primary measure called the expected average overlap (EAO).

Results on VOT2020 are shown in Table~\ref{tab:vot20}. The proposed SAM2.1++ outperforms all compared trackers. 
In particular, it outperforms the recently published MixViT~\cite{mixvit_tpami2024} by 25\% EAO, improving both, accuracy and robustness. 
SAM2.1++ outperforms also the VOT2020 challenge winner RPT~\cite{rpt_eccvw2020} by a large margin (37.5\% in EAO). 
Comparing SAM2.1++ to the original SAM2.1, the EAO is boosted by 7\%, while accuracy and robustness are improved by 2.7\% and 2.1\%, respectively. 

\begin{table}[t]
\centering
\caption{State-of-the-art comparison on the VOT2020 benchmark. The challenge winner is marked by~\includegraphics[height=0.7em]{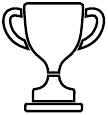}.
}
\label{tab:vot20}
\begin{tabular}{llll}
\toprule
  & EAO & Accuracy & Robustness \\
\midrule
ODTrack \small{~\cite{odtrack}}  & 0.605 \third{} & 0.761 & 0.902 \third{} \\
MixViT-L+AR \small{~\cite{mixvit_tpami2024}} & 0.584 & 0.755 & 0.890  \\
SeqTrack-L \small{~\cite{seqtrack}} & 0.561 & - & -  \\
MixFormer-L \small{~\cite{mixformer_cvpr2022}} & 0.555 & 0.762 \third{} & 0.855  \\
RPT~\includegraphics[height=0.7em]{trophy_e.pdf} {\small{~\cite{rpt_eccvw2020}}}  & 0.530 & 0.700 & 0.869  \\
OceanPlus \small{~\cite{ocean}}  & 0.491 & 0.685 & 0.842  \\
AlphaRef \small{~\cite{alpharef_cvpr2021}}   & 0.482 & 0.754 & 0.777  \\
AFOD  {\small{~\cite{afod}}}     & 0.472 & 0.713 & 0.795  \\
\midrule
SAM2.1 \small{~\cite{sam2}}       & 0.681 \second{} & 0.778 \second{} & 0.941 \second{} \\
SAM2.1++    & {\bf 0.729} \first{} & {\bf 0.799} \first{} & {\bf 0.961} \first{} \\
\bottomrule
\end{tabular}
\end{table}

{\bf The VOT2022 benchmark}~\cite{vot2022} uses a refreshed dataset with 62 sequences (simplest sequences removed, and more challenging added).  
Table~\ref{tab:vot22} includes the challenge top-performers, including the winner MS\_AOT~\cite{ms_aot}, as well as the recent published state-of-the-art trackers: DiffusionTrack~\cite{diffusiontrack}, MixFormer~\cite{mixformer_cvpr2022}, OSTrack~\cite{ostrack_eccv2022} and D3Sv2~\cite{d3s_tpami}. 
The proposed SAM2.1++ outperforms the VOT2022 winner MS\_AOT by a significant margin of 12\% in EAO. 
Note that the performance improvement is a consequence of both improved accuracy (2\%) and robustness (3\%) compared to the MS\_AOT. 
In addition to achieving state-of-the-art performance, SAM2.1++ outperforms also the baseline SAM2.1 by a healthy margin of 9\% EAO. 

The results on both, VOT2020 and VOT2022 clearly show that SAM2.1++
outperforms all trackers, including the challenges top-performers and the recently published trackers, setting new state-of-the-art on these benchmarks. 
Despite its simplicity, the proposed {\it training-free} memory management is the key element for achieving excellent tracking performance.

\begin{table}[ht]
\centering
\caption{State-of-the-art comparison on the VOT2022 benchmark. The challenge winner is marked by~\includegraphics[height=0.7em]{trophy_e.pdf}.
}
\label{tab:vot22}
\begin{tabular}{llll}
\toprule
  & EAO & Accuracy & Robustness \\
\midrule
MS\_AOT~\includegraphics[height=0.7em]{trophy_e.pdf} {\small ~\cite{ms_aot}} & 0.673 \third{} & 0.781 & 0.944 \third{}  \\
DiffusionTrack \small{~\cite{diffusiontrack}} & 0.634 & - & -  \\
DAMTMask  \small{~\cite{vot2022}}     & 0.624 & 0.796 \third{} & 0.891  \\
MixFormerM \small{~\cite{mixformer_cvpr2022}}   & 0.589 & 0.799 \second{} & 0.878  \\
OSTrackSTS \small{~\cite{ostrack_eccv2022}}     & 0.581 & 0.775 & 0.867  \\
Linker \small{~\cite{linker_ieee2024}}         & 0.559 & 0.772 & 0.861  \\
SRATransTS \small{~\cite{vot2022}}     & 0.547 & 0.743 & 0.866  \\
TransT\_M  \small{~\cite{transt_cvpr2021}}    & 0.542 & 0.743 & 0.865  \\
GDFormer \small{~\cite{vot2022}}      & 0.538 & 0.744 & 0.861  \\
TransLL \small{~\cite{vot2022}}        & 0.530 & 0.735 & 0.861  \\
LWL\_B2S \small{~\cite{lwl_eccv2020}}      & 0.516 & 0.736 & 0.831  \\
D3Sv2 \small{~\cite{d3s_tpami}}         & 0.497 & 0.713 & 0.827  \\
\midrule
SAM2.1 \small{~\cite{sam2}}        & 0.692 \second{} & 0.779 & 0.946 \second{} \\
SAM2.1++       & {\bf 0.753} \first{} & {\bf 0.800} \first{} & {\bf 0.969} \first{} \\
\bottomrule
\end{tabular}
\end{table}

{\bf VOTS2024 benchmark.} For a complete evaluation on VOT, we report the performance also on the most recent VOTS2024 benchmark.
In contrast to VOT2020 and VOT2022, the VOTS2024 benchmark introduces a new, larger dataset, tracking multiple objects in the same scene (with ground truth sequestered at an evaluation server) and a new performance measure, designed to address short-term, long-term\footnote{The target may leave the field of view and re-appear later in the video.}, single and multi-target tracking scenarios. 
The VOTS2024 is currently considered as the most challenging tracking benchmark. 

Results are reported in Table~\ref{tab:vot24}.
It is worth pointing out that the top performers are mostly unpublished (not peer-reviewed) trackers, tuned for the competition, and often complex ad-hoc combinations of multiple methods. 
For example, the challenge winner S3-Track combines visual and (mono)depth features, uses several huge backbones, and is much more complex than SAM2.1++.
Despite this, using the same parameters as in other experiments, SAM2.1++ achieves a solid second place among the challenging VOTS2024 competition.
In particular, it outperforms the challenge-tuned versions of the recently published trackers LORAT~\cite{lorat_eccv2024}, Cutie~\cite{cutie_cvpr2024}, and the VOT2022 winner AOT~\cite{ms_aot}. 
Furthermore, the proposed memory management mechanism in SAM2.1++ contributes to 8\% performance boost in tracking quality compared to the baseline SAM2.1, mostly due to 9\% higher robustness. 

\begin{table}[th]
\centering
\caption{State-of-the-art comparison on the VOT2024 benchmark. The challenge winner is marked by~\includegraphics[height=0.7em]{trophy_e.pdf}.}
\label{tab:vot24}
\begin{tabular}{llll}
\toprule
  & Quality & Accuracy & Robustness \\
\midrule
S3-Track~\includegraphics[height=0.7em]{trophy_e.pdf} \small{~\cite{s3_arxiv}}  & {\bf 0.722} \first{} & 0.784 & {\bf 0.889} \first{}  \\
DMAOT\_SAM\small{~\cite{deaot_neurips2022}} & 0.653 & {\bf 0.794} \first{} & 0.780  \\
HQ-DMAOT\small{~\cite{deaot_neurips2022}}  & 0.639 & 0.754 & 0.790  \\
DMAOT\small{~\cite{deaot_neurips2022}}     & 0.636 & 0.751 & 0.795 \third{}  \\
LY-SAM \small{~\cite{vots2024}}  & 0.631 & 0.765 & 0.776  \\
Cutie-SAM \small{~\cite{cutie_cvpr2024}} & 0.607 & 0.756 & 0.730  \\
AOT   \small{~\cite{ms_aot}}     & 0.550 & 0.698 & 0.767  \\
LORAT \small{~\cite{lorat_eccv2024}}      & 0.536 & 0.725 & 0.784  \\
\midrule
SAM2.1 \small{~\cite{sam2}}   & 0.661 \third{} & 0.791 \third{} & 0.790 \\
SAM2.1++  & 0.711 \second{} & 0.793 \second{} & 0.864 \second{} \\
\bottomrule
\end{tabular}
\end{table}

%-------------------------------------------------------------------------
\subsection{SoTa comparison on bounding box benchmarks}  \label{sec:exp-bb}

For a complete evaluation, we compare SAM2.1++ on the following three standard bounding box tracking datasets: LaSoT~\cite{lasot_cvpr19}, LaSoT$_{\mathrm{ext}}$~\cite{lasot_ijcv} and GoT10k~\cite{got10k}. 
Since frames are annotated by bounding boxes and SAM2 requires a segmentation mask provided in the first frame, we use the same SAM2 model to estimate the initialization mask. 
The min-max operation is applied on the predicted masks to obtain the axis-aligned bounding boxes required for the evaluation. 
Tracking performance is computed using area under the success rate curve~\cite{otb_pami2015} (AUC) in LaSoT~\cite{lasot_cvpr19} and LaSoT$_{\mathrm{ext}}$~\cite{lasot_ijcv} and the average overlap~\cite{got10k} (AO) in Got10k. 

{\bf LaSoT}~\cite{lasot_cvpr19} is a large-scale tracking dataset with 1,400 video sequences, with 280 evaluation sequences and the rest are used for training. 
The sequences are equally split into 70 categories, where each category is represented by 20 sequences (16 for training and 4 for evaluation). 
The dataset consists of various scenarios covering short-term and long-term tracking.
Results are shown in Table~\ref{tab:bbox}. 
The proposed SAM2.1++ outperforms the baseline SAM2.1~\cite{sam2} by 7.3\%, which indicates that the proposed memory management is important also in a bounding box tracking setup. 
Furthermore, SAM2.1++ performs on par with the top-performing tracker LORAT~\cite{lorat_eccv2024}, which was tuned on LaSoT training set, i.e., on the categories included in the evaluation set. 
It is worth noting that LORAT~\cite{lorat_eccv2024} has approximately 50\% more training parameters than SAM2.1++, making the model significantly more complex. 

{\bf LaSoT$_{\mathrm{ext}}$}~\cite{lasot_ijcv} is an extension of the LaSoT~\cite{lasot_cvpr19} dataset, by 150 test sequences, divided into 15 new categories, which are not present in the training dataset. 
The results in Table~\ref{tab:bbox} show that SAM2.1++ outperforms the baseline version by a comfortable margin of 7\% in AUC. 
In addition, it outperforms the second-best tracker LORAT~\cite{lorat_eccv2024} by 7.6\%. 
This indicates that SAM2.1++ generalizes well across various object categories while existing trackers suffer a much larger performance drop.

{\bf GoT10k}~\cite{got10k} is another widely used large-scale tracking dataset, composed of approximately 10k video sequences, from which 180 sequences are used for tracking evaluation. 
We observed that top-performing trackers on GoT10k test set achieve excellent tracking performance, e.g., more than 78\% of average overlap, which leaves only small room for potential improvements. 
However, a solid 3.7\% boost in tracking performance is observed when comparing SAM2.1++ to the top-performers LORAT~\cite{lorat_eccv2024} and ODTrack~\cite{odtrack}. 
A close inspection of the SAM2.1++ results reveals that more than 99\% of frames are successfully tracked (i.e., with a non-zero overlap), which indicates that GoT10k ~\cite{got10k} difficulty level is indeed diminishing for modern trackers.

\begin{table}[ht]
\centering
\caption{State-of-the-art comparison on three standard bounding-box benchmarks.}
\label{tab:bbox}
\begin{tabular}{llll}
\toprule
  & LaSoT & LaSoT$_{\mathrm{ext}}$ & GoT10k \\
  & (AUC) & (AUC) & (AO)  \\
\midrule
MixViT \small{~\cite{mixvit_tpami2024}}     & 72.4 & -    & 75.7 \\
LORAT \small{~\cite{lorat_eccv2024}}      & {\bf 75.1} \first{} & 56.6 \third{} & 78.2 \third{} \\
ODTrack \small{~\cite{odtrack}}    & 74.0 \second{} & 53.9 & 78.2 \third{} \\
DiffusionTrack \small{~\cite{diffusiontrack}} & 72.3 &  -  & 74.7 \\
DropTrack \small{~\cite{droptrack_cvpr2023}}    & 71.8 & 52.7 & 75.9 \\
SeqTrack \small{~\cite{seqtrack}}      & 72.5 \third{} & 50.7 & 74.8 \\
MixFormer \small{~\cite{mixformer_cvpr2022}} & 70.1 &  -  & 71.2 \\
GRM-256 \small{~\cite{grm-256_cvpr2023}}      & 69.9 &  -   & 73.4 \\
ROMTrack \small{~\cite{romtrack_iccv2023}}     & 71.4 & 51.3 & 74.2 \\
OSTrack \small{~\cite{ostrack_eccv2022} }     & 71.1 & 50.5 & 73.7 \\
KeepTrack \small{~\cite{keeptrack}}    & 67.1 &  48.2   & -    \\
TOMP \small{~\cite{tomp_cvpr2022}}         & 68.5 &  -   & -    \\
\midrule
SAM2.1 \small{~\cite{sam2}}       & 70.0 & 56.9 \second{} & 80.7 \second{} \\
SAM2.1++      & {\bf 75.1} \first{} & {\bf 60.9} \first{} & {\bf 81.1} \first{} \\
\bottomrule
\end{tabular}
\end{table}

\section{Conclusion}  \label{sec:conclusion}

We proposed a new distractor-aware memory model DAM and a management regime for memory-based trackers. 
The new model divides the memory by its tracking functionality into the recent appearances memory (RAM) and a distractor-resolving memory (DRM), responsible for the tracking accuracy and robustness, respectively. 
Efficient update rules are proposed, which also utilize the tracker output to detect frames with critical distractors useful for updating DRM. 
In addition, a distractor-distilled dataset DiDi is proposed to facilitate studying tracking in challenging \mbox{scenarios}.

The proposed DAM memory is implemented with SAM2.1~\cite{sam2}, forming SAM2.1++. Extensive analysis confirms the design decisions. 
Without any retraining and using fixed parameters, SAM2.1++ sets a solid state-of-the art on six benchmarks with a moderate (20\%) speed reduction compared to SAM2.1 (i.e., 11 fps vs 13.3 fps).
This makes a compelling case for arguably simpler localization architectures of memory-based frameworks compared to the current tracking state-of-the-art. 
Furthermore, the results suggest more research should focus on efficient memory designs, with possibly learnable management policies. We believe these directions hold a strong potential for further performance boosts in future work.

\textbf{Acknowledgements.}
This work was supported by Slovenian research agency program P2-0214 and projects 
J2-2506, % Davimar
L2-3169, % MV4.0
Z2-4459  % Transtrack
and COMET, and by supercomputing network SLING (ARNES, EuroHPC Vega - IZUM). 

{
    \small
    \bibliographystyle{ieeenat_fullname}
    \bibliography{main}
}

\newpage \clearpage

\section{Appendix}  \label{sec:appendix}

%----------------------------------------------------------------
\subsection{Impact of the model size}  \label{sec:abl-model-size}

The segment anything model 2 (SAM2)~\cite{sam2} was originally developed in four model sizes, denoted by tiny (T), small (S), base (B) and large (L). 
In Table~\ref{tab:model_size} these four model sizes of unchanged SAM2.1 are compared with our SAM2.1++ version, presented in the paper on the new DiDi dataset. 
Results show a clear and consistent performance improvement across all four model sizes. 
In particular, the tracking quality improves by approximately  6\% or 7\%, depending on the model size, mostly due to the improved robustness. 
These results show 
that the proposed distractor-aware memory generalizes well over various model sizes, demonstrating that the model has not been tuned to the exact SAM2 model. 

\begin{table}[h]
\centering
\caption{Comparison of different model sizes on DiDi dataset. The ++ denotes the proposed tracker, while the T, S, B, L denote the tiny, small, base and large Hiera backbone sizes, respectively. {\it Params} denotes number of parameters, while {\it Acc.} and {\it Rob.} denote accuracy and robustness, respectively.}
\label{tab:model_size}
\begin{tabular}{lllll}
\toprule
  & {\bf Params} & {\bf Quality} & {\bf Acc.} & {\bf Rob.} \\
\midrule
SAM2.1-T   & 39M & 0.600 & 0.697 & 0.848 \\
\rowcolor{lgray} SAM2.1++-T  & 39M & 0.642 $\uparrow$7\% & 0.695 & 0.907 \\
SAM2.1-S    & 46M & 0.630 & 0.718 & 0.866 \\
\rowcolor{lgray} SAM2.1++-S   & 46M & 0.668 $\uparrow$6\% & 0.709 & 0.930 \\
SAM2.1-B    & 81M & 0.624 & 0.721 & 0.856 \\
\rowcolor{lgray} SAM2.1++-B  & 81M & 0.664 $\uparrow$6\% & 0.709 & 0.930 \\
SAM2.1-L    & 224M & 0.649 & 0.720 & 0.887 \\
\rowcolor{lgray} SAM2.1++-L  & 224M & 0.694 $\uparrow$7\% & 0.727 & 0.944 \\
\bottomrule
\end{tabular}
\end{table}

%----------------------------------------------------------------
\subsection{Impact of the model version}  \label{sec:abl-model-version}

This section compares the performance improvements for the two individual SAM versions, i.e., SAM2 and SAM2.1. The SAM2.1 version improves the initial version in handling small and visually similar objects by introducing additional augmentation techniques in training. 
It also includes improved occlusion handling by training the model on longer frame sequences. 

Results are shown in  Table~\ref{tab:model_version}.
To demonstrate the improvement of the 2.1 model version over version 2, we compare the SAM2 with the SAM2.1 on the DiDi dataset, which results in approximately 3.5\% improvement in tracking quality. 
Next, we compare the original SAM2 and the version with our new memory model (i.e., SAM2++). The tracking performance improves by 7\%, which is well beyond the performance improvement from SAM2 to SAM2.1 and supports the importance of a high-quality memory model and the memory management regime. 
A simimlar performance boost (close to 7\%) is observed between SAM2.1 and SAM2.1++, which implies complementarity of the new memory model with the baseline method performance imporvements that come from better training.
Similarly as in Section~\ref{sec:abl-model-size}, we conclude that the proposed distractor-aware memory is robust to different model versions, demonstrating a consistent improvements in tracking performance on two SAM2 versions. 

\begin{table}[h]
\centering
\caption{Comparison of two SAM model versions: SAM2 and SAM2.1 on DiDi dataset.}
\label{tab:model_version}
\begin{tabular}{llll}
\toprule
  & Quality & Accuracy & Robustness \\
\midrule
SAM2      & 0.627 & 0.723 & 0.850 \\
\rowcolor{lgray} SAM2++    & 0.668 $\uparrow$7\% & 0.710 & 0.929 \\
SAM2.1    & 0.649 & 0.720 & 0.887 \\
\rowcolor{lgray} SAM2.1++  & 0.694 $\uparrow$7\% & 0.727 & 0.944 \\
\bottomrule
\end{tabular}
\end{table}

%----------------------------------------------------------------
\subsection{Real-time performance}  \label{sec:real-time}

We further evaluate the proposed SAM2.1++ under real-time tracking constraints. We thus evaluate the tracker on VOT2022-RT~\cite{vot2022} challenge\footnote{SAM2.1 and SAM2.1++ were evaluated on the machine with the AMD EPYC 7763 64-Core 2.45 GHz CPU and Nvidia A100 40GB GPU.}, which was specifically designed for real-time evaluation.
Specifically, VOT challenges are run by VOT toolkits, which manage the real-time constraints. A frame is sent to the tracker, which needs to process it and report the target position at 20FPS frame rate.
If the tracker is not able to process the frame in time, the prediction from the previous frame is used as the estimate for the current frame and next frame is sent to the tracker. 
Such a setup simulates actual real-time scenario, which is much more realistic than reporting just the average tracking speed. The tracking performance is measured using standard VOT2022 measures~\cite{vot2022}: the primary measure expected average overlap (EAO), and two auxiliary measures, i.e., accuracy and robustness. 

The results in Table~\ref{tab:real_time} show that the proposed SAM2.1++ (L model size) outperforms all trackers that participated in VOT2022-RT challenge. 
In particular, it outperforms the challenge winner by 4\% in EAO demonstrating excellent real-time performance. 
These results show that the proposed distractor-aware memory adds only a small computational overhead, yet bringing remarkable robustness capabilities and making it useful for real applications.

\begin{table}[h]
\centering
\caption{Real-time performance on VOT2022-RT challenge. The challenge winner is marked by~\includegraphics[height=0.7em]{trophy_e.pdf}.}
\label{tab:real_time}
\begin{tabular}{llll}
\toprule
  & EAO & Accuracy & Robustness \\
\midrule
\rowcolor{lgray} MS\_AOT~\includegraphics[height=0.7em]{trophy_e.pdf}    & 0.610 \third{} & 0.751 \second{} & 0.921 \third{}  \\
OSTrackSTS & 0.569 & 0.766 \first{} & 0.860  \\
\rowcolor{lgray} SRATransTS & 0.547 & 0.743 \third{} & 0.866  \\
\midrule
SAM2.1    & 0.614 \second{} & 0.722 & 0.922 \second{} \\
\rowcolor{lgray} SAM2.1++  & 0.635 \first{} & 0.717 & 0.942 \first{} \\
\bottomrule
\end{tabular}
\vspace{-0.2cm}
\end{table}

%----------------------------------------------------------------
\subsection{Sensitivity to threshold values}

We analyze the sensitivity of the proposed SAM2.1++ to the exact value of the manually determined parameters. 
In particular, we focus on three  thresholds defined in {\it Distractor-resolving memory} (Section~3.2.2). 
Experiments were conducted on the VOT2022~\cite{vot2022} dataset using the unsupervised experiment to ensure fast execution and compute the average overlap (AO) as the performance measure. 

\begin{figure}[ht]
  \centering
  \includegraphics[width=0.8\linewidth]{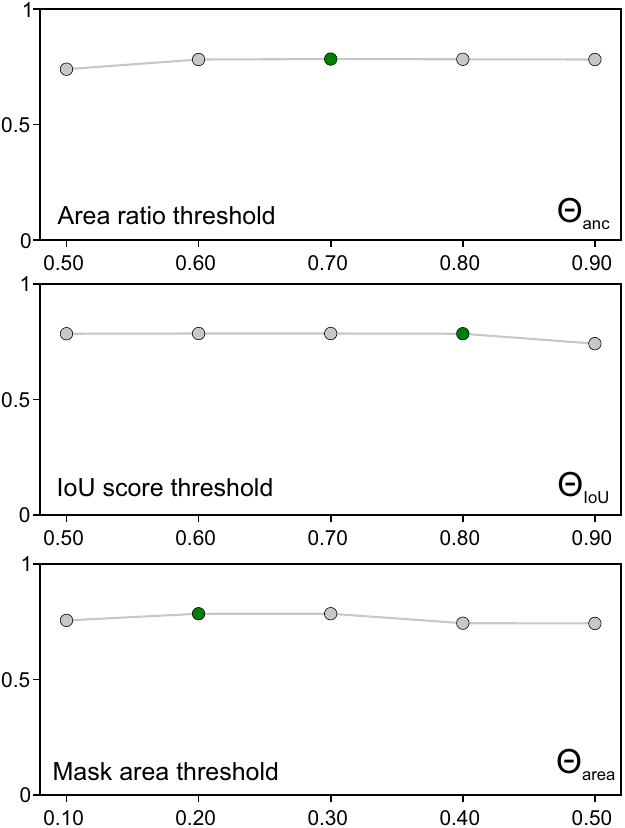} 
  \caption{Sensitivity of the SAM2.1++ to different values of three thresholds: $\Theta_{anc}$, $\Theta_{IoU}$ and $\Theta_{area}$. Experiments were done on VOT2022 using average overlap as the performance measure. The selected threshold value is marked by a green circle. }
\label{fig:suppl-parameters}
\vspace{-0.2cm}
\end{figure}

We analyzed the influence of the threshold for the ratio between the target and alternative predicted masks, i.e. $\Theta_{anc}$ = 0.7. 
This ratio is used for preemptive update upon distractor detection  and is thus crucial for our distractor-aware memory. 
The results in Figure~\ref{fig:suppl-parameters} show that tracking results are extremely stable for a wide range of $\Theta_{anc} \in [0.6, 0.9]$ demonstrating the robust design of the tracker. 

Next, we tested the IoU score threshold ($\Theta_{Iou}$ = 0.8), used to determine if the predicted mask is reliable for distractor testing. 
Results in Figure~\ref{fig:suppl-parameters} show that tracking performance is extremely stable for a wide range of parameters ($\Theta_{IoU} \in [0.5, 0.8]$) scoring almost identical AO. 

Finally, the mask area threshold was tested, i.e. $\Theta_{area}$ = 0.2. 
This threshold is used to determine the tracking stability and is together with the $\Theta_{IoU}$ a necessary condition to trigger the DRM update.
Results in Figure~\ref{fig:suppl-parameters} show stable tracking performance across the wide range of thresholds and confirms that SAM2.1++ is not sensitive to the exact value of this parameter. 

%----------------------------------------------------------------
\subsection{DiDi dataset statistics}

In this section we provide additional information about the distractor-distilled dataset DiDi construction. 
In particular, number of sequences from each source dataset is given in the following: 
\begin{itemize}
    \item LaSoT~\cite{lasot_cvpr19}: 86 sequences
    \item UTB-180~\cite{utb180}: 56 sequences
    \item VOT2022-ST~\cite{vot2022}: 20 sequences
    \item VOT2022-LT~\cite{vot2022}: 7 sequences
    \item VOT2020-LT~\cite{kristan_vot2020}: 6 sequences
    \item GoT10k~\cite{got10k}: 4 sequences
    \item VOT2020-ST~\cite{kristan_vot2020}: 1 sequences \\
    \noindent\rule{7cm}{0.4pt}
    \item Total: 180 sequences (274,882 frames)
\end{itemize}

%----------------------------------------------------------------
\subsection{Qualitative analysis}

\begin{figure*}[t]
  \centering
  \includegraphics[width=\linewidth]{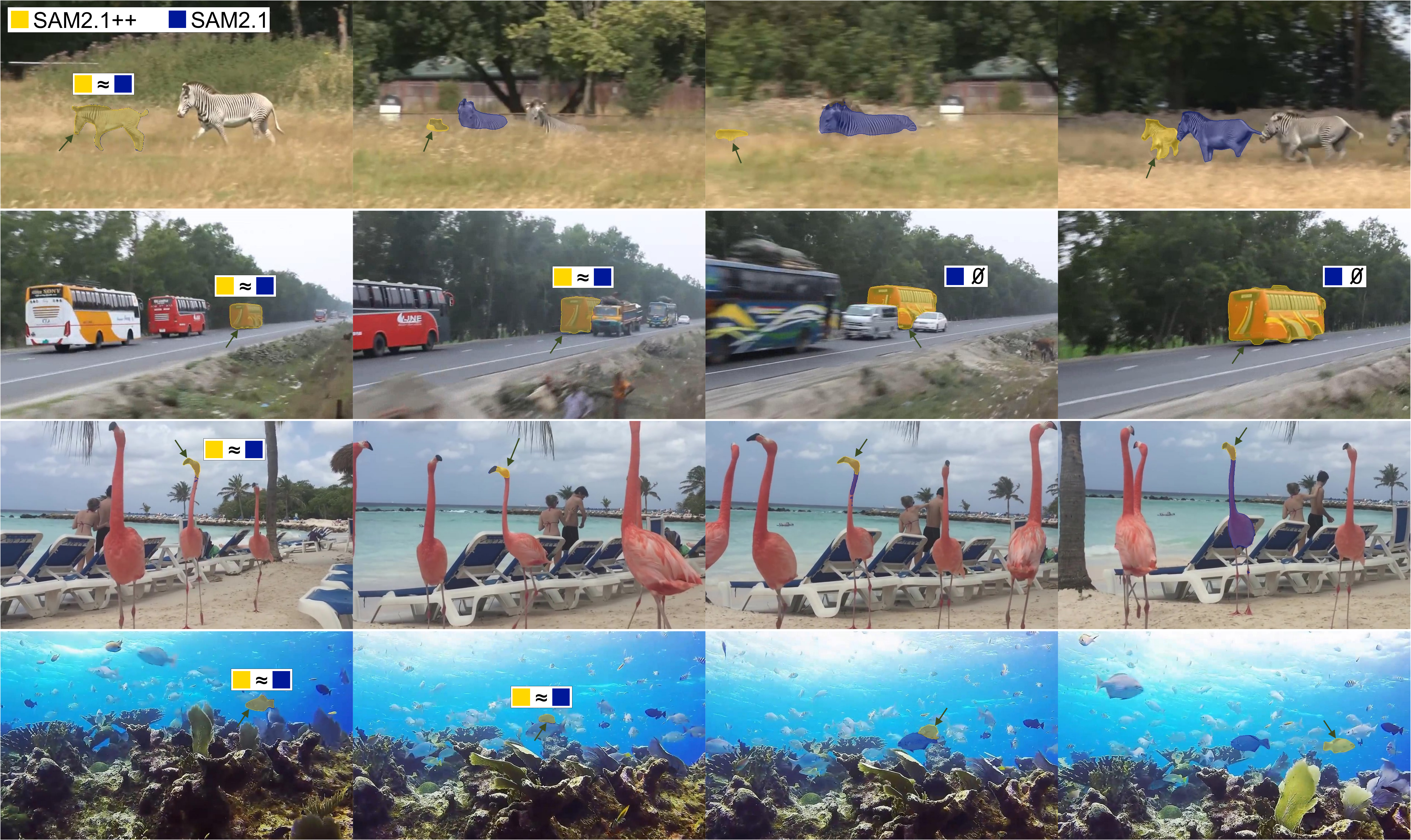} 
  \caption{Qualitative comparison of the baseline SAM2.1 (blue) and the proposed SAM2.1++ (yellow). The symbol $\approx$ denotes approximately identical outputs and $\emptyset$ denotes an empty prediction (i.e, mask with all-zeros). Tracked object is denoted with a green arrow.}
\label{fig:qualitative_suppl}
\end{figure*}

Figure~\ref{fig:qualitative_suppl} presents a qualitative comparison between the baseline SAM2.1 and the proposed SAM2.1++ on four video sequences.
In the first row a zebra is tracked with other zebras in its vicinity. 
When the zebra is partially occluded, SAM2.1 drifts to the wrong zebra and starts to track it, while SAM2.1++ tracks only the visible part of the target during occlusion and stays on the selected zebra until the end of the sequence. 

In the second row of Figure~\ref{fig:qualitative_suppl}, the baseline SAM2.1 tracker successfully tracks the bus until the full occlusion and fails to re-detect it after the re-appearance. This failure occurs due to the too frequent memory updates when target is occluded and is successfully addressed with the proposed memory update in SAM2.1++. 

The third row in Figure~\ref{fig:qualitative_suppl} shows tracking  of a flamingo's head. 
The baseline SAM2.1 tends to jump on the bird's beak or extend to the whole body, since it prefers to segment the regions with so-called high objectness (i.e., regions with well-defined edges). 
The proposed SAM2.1++ successfully tracks the flamingo's head even if the edge between the head and the neck is not clearly visible. 
In this case, part of the neck is segmented by an alternative mask and thus detected as a distractor. 
Updating the distractor resolving memory (DRM) using such {\it critical} frames results in a more stable and accurate tracking. 

A similar effect is demonstrated in the fourth row of Figure~\ref{fig:qualitative_suppl}, where a fish similar to the tracked fish occludes it and causes SAM2.1 to jump to it. 
On the other hand, SAM2.1++ successfully detects such critical frames, updates the DRM and avoids the tracking failure. 

\end{document}